# Physics-Distilled Neural Network enabled by Large Language Models for Manufacturing Process-Property Predictive Modeling

**Ge Song[1], Kiarash Naghavi Khanghah[1], Anandkumar Patel[2], Rajiv Malhotra[2], Hongyi Xu[1]***

[1] School of Mechanical, Aerospace and Manufacturing Engineering, University of Connecticut, Storrs, CT 06269

[2] Department of Mechanical & Aerospace Engineering, Rutgers, the State University of New Jersey, Piscataway, NJ 08854

* Email: hongyi.3.xu@uconn.edu

## Abstract

Predicting process-property relationships in manufacturing is often challenged by high experimental costs and the limited interpretability of complex 'black-box' models. This paper proposes a novel knowledge distillation framework designed to achieve high-accuracy predictions in data-scarce scenarios. The framework integrates analytical physics priors, which are systematically extracted from scientific literature via Large Language Models, into a privileged teacher model. We employ a Graph-Masked Attention layer to capture the complex physical dependencies among input variables showing strict setpoints or a combination of static and high-frequency temporal signatures. This privileged knowledge is distilled into a lightweight student predictor for inference. The feasibility and robustness of the framework are evaluated through a comprehensive experiment across five diverse manufacturing processes. To ensure statistical reliability, given the small dataset sizes, a repeated K-fold cross-validation technique is employed to quantify model stability and generalization. Results indicate that the proposed framework consistently achieves high predictive accuracy across all evaluated domains. Most importantly, the architecture demonstrates significant fault tolerance by maintaining robust predictive performance even in scenarios where LLM-derived analytical priors are suboptimal or incomplete. Furthermore, the student predictor achieves an inference frequency exceeding 6000 Hz, which facilitates real-time edge deployment on standard industrial hardware. This work provides a scalable solution for bridging the gap between theoretical physics and real-time industrial monitoring in data-limited environments.

## 1. Introduction

Modern advanced manufacturing, ranging from precision laser-based material processing [1] to additive fabrication [2], is fundamentally constrained by the challenge of process control [3]. An accurate understanding of the complex relationship between process parameters and predictive outcomes is generally required to achieve high geometric fidelity and structural integrity [4] [5]. Moreover, the inability to accurately predict process outcomes imposes a severe economic burden on high-precision manufacturing systems [6]. When predictive models fail, industries are forced to rely solely on iterative 'trial-and-error' experimentation, which leads to significant material waste, increased energy consumption, and long product development cycles [7] [8].

Predictive modeling of process-property relationships, however, presents a significant bottleneck. Physics-based models typically become unreliable when addressing multi-physics interactions [9] [10] [11], while standard data-driven approaches require large experimental datasets for effective training [12]. In practice, generating large experimental datasets is prohibitively expensive and time-consuming, which routinely results in data scarcity [13]. When trained on such sparse data, conventional machine learning models are often non-interpretable by failing to provide scientific insight [14], and generalizing poorly to unseen process conditions [15].

One emerging strategy to mitigate data scarcity is to leverage the rich prior knowledge embedded within Large Language Models (LLMs) that have been pre-trained and tuned on vast repositories of scientific and technical literature [16] [17]. Retrieval-Augmented Generation (RAG) [18] [19] [20] enables LLMs to ground their outputs in domain-specific corpora rather than relying solely on parametric memory, which has been shown to improve factual consistency and reduce hallucination in technical question answering [19] [20]. Within manufacturing, a growing amount of work explores LLMs both as conversational knowledge assistants and as quantitative modeling agents [21] [22]. Liu et al. [23] used LLMs with RAG to perform named entity recognition for additive manufacturing process descriptions, enabling automated taxonomy construction. AMGPT [24] retrieves from a curated additive manufacturing literature base to answer contextual queries on metal printing, while related ontology-based [25] and knowledge-graph-augmented frameworks [26] have been proposed for process planning and design for AM tasks. On the predictive side, AdditiveLLM [27] fine-tunes LLMs to classify keyholing, lack-of-fusion, and balling defect regimes in laser powder bed fusion directly from process parameters, and Cao et al. [28] integrate LLM-based parameter advisors with a stacking neural model for binder jetting of Ti-6A1-4V.

A particularly relevant direction couples LLMs with symbolic regression to extract analytical relationships, often informed by the broader literature on LLM-driven equation discovery [29] [30]. Our previous work [9] established an LLM-based framework that integrates automated literature retrieval with iterative model refinement to discover analytical parametric relationships, and demonstrated strong extrapolative performance across machining, deformation, and additive processes with limited experimental data. However, this work and related approaches share three limitations relevant to the present work. First, the generated relationships are rigid analytical functions whose structural form is constrained by what can be expressed in compact symbolic terms, and which therefore struggle to capture complex nonlinear interactions outside the scope of simplified theoretical expressions. Second, the quality of the functions is limited by the breadth and accuracy of the available literature, where sparse or inconsistent sources yield misleading physical priors. Third, iterative refinement is computationally expensive and offers no architectural mechanism to recover when the prior remains suboptimal. These observations motivate using the LLM as a generator of analytical priors rather than as the final predictor, and embedding those priors into a more flexible neural architecture that can compensate when the priors are imperfect.

Physics-incorporated neural networks [31] address data scarcity from the opposite direction: rather than retrieving knowledge from text, they encode guiding equations into the learning process itself. In a Physics-Informed Neural Network (PINN), the residuals of partial differential equations and boundary conditions are added to the loss function, computed via automatic differentiation, which enforces physically consistent predictions and reduces dependence on large datasets [32]. Manufacturing applications have developed across thermal and structural domains. Chen et al. [33] proposed a physics-guided neural operator for composites manufacturing, and Wang et al. [34] embedded a physics-guided constraint for tool-wear prediction in machining. Filipovic et al. [35] developed a physics-guided feedforward controller for seamless pipe manufacturing, while Wang et al. [36] embedded physical knowledge to predict phonon dispersion in cellular metamaterials. In injection molding, Pieressa et al. [37] used PINNs to predict weld-line visibility and in directed energy deposition. Peng et al. [38] used a transfer-learning PINN to reconstruct three-dimensional temperature fields from a single two-dimensional measurement. Recent reviews [12] [39] [40] survey the broader landscape and consistently identify physics-informed methods as a path toward physically consistent and data-efficient manufacturing models. Despite this progress, physics-incorporated neural networks share a structural limitation that has received less attention: the physical loss term and the guiding equations always involve human experts. Many manufacturing processes, including emerging additive and hybrid processes, lack closed-form equations that map directly onto controllable process parameters. In these cases, deriving an appropriate physical loss requires extensive domain expertise, manual literature synthesis, and iterative refinement of the constraint formulation. This bottleneck limits the feasibility of PINN-style approaches to the diverse and rapidly evolving landscape of advanced manufacturing.

Knowledge distillation [41] is a training paradigm in which a complex teacher model transfers its learned representations and decision boundaries to a lightweight student model, originally motivated by model compression for resource-constrained deployment. In manufacturing, knowledge distillation has been adopted primarily in classification-oriented tasks, where it enables real-time fault and defect detection on edge hardware. A recent systematic review [42] documents its widespread use across predictive maintenance, including bearing fault diagnosis on FPGA hardware, gear fault classification with relational distillation, and milling-cutter monitoring using vision transformers as teachers and lightweight transformer students [43]. For decentralized manufacturing, Liu et al. [44] used distillation to share monitoring knowledge across geographically separated production units. Closer to the present work, Liu et al. [45] used knowledge distillation to automatically discover the network structure of PINNs, including that distillation can transfer physics-aware representations and not only statistical ones. Tang et al. [46] combined PINN-accelerated mechanistic modeling with knowledge distillation for real-time predictive control of biopharmaceutical manufacturing. Beyond classical distillation, the Learning Using Privileged Information (LUPI) paradigm introduced by Vapnik and Vashist [47] uses privileged features available at training but absent at inference to guide the student, providing a principled framework when training-time information is richer than runtime information. Although the knowledge distillation is well established for classification in manufacturing, its use for continuous process-property regression remains comparatively limited. In addition, prior manufacturing distillation work seldom integrates an explicit physics-derived inductive bias into the teacher where, instead, the teacher is typically a larger black-box predictor.

To address the constraints identified above, we propose a framework that integrates automatically extracted analytical equations by LLMs into a distillation-based neural network architecture. We introduce a hybrid physics-distilled neural network with a *Privileged Teacher* model that utilizes a Graph-Masked Attention (GMA) layer to structurally formulate latent representations within physical trends and privileged training features. This knowledge is distilled into a lightweight *Student Predictor* for fast outcome prediction. This design can predict complex process-property relationships in a manner that significantly reduces the reliance on costly experimentation. The novelties of this work include:

- **Physics-Distilled Teacher-Student Architecture:** The paper introduces a novel framework that integrates previously established LLM-derived analytical physics priors with privileged knowledge to solve the structural rigidity of standalone analytical models.
- **Physics-Constrained Graph-Masked Attention (GMA):** A Privileged Teacher architecture employs a specialized GMA layer to structurally enforce physical partial derivatives, which prevents the model from learning non-physical feature correlations.
- **Asymmetric Latent-Manifold Distillation:** By transferring physics-guided latent representations from the complex teacher to a lightweight Student Predictor, the architecture achieves high-fidelity predictions relying exclusively on standard machine setting parameters during runtime.
- **Fault Tolerance and Edge-Deployment Scalability:** The architecture maintains robust prediction accuracy even when the analytical priors are suboptimal or incomplete, while achieving fast inference speeds exceeding 6000 Hz to enable real-time industrial monitoring.

The remainder of this paper is organized as follows: Section 2 details the proposed methodology; Section 3 presents experimental results and discussion; and Section 4 concludes the study.

## 2. METHODOLOGY

The proposed framework, as illustrated in Figure 1, integrates Large Language Models with a distillation-based neural network architecture to enable efficient prediction of manufacturing process-property relationships. This framework consists of three primary components: automated knowledge extraction, physics-incorporated teacher modeling, and latent-space knowledge distillation within the student predictor.

First, a Large Language Model acts as a knowledge-extraction engine to generate analytical physics priors directly from research literature. These priors are then embedded into a teacher model with privileged information via a Graph-Masked Attention layer. This layer structurally guides the model by incorporating physical partial derivatives and privileged features into its latent space. Finally, through a knowledge distillation process, the physical features learned by the teacher are transferred to a lightweight predictor to enable robust and computationally efficient prediction even with data scarcity.

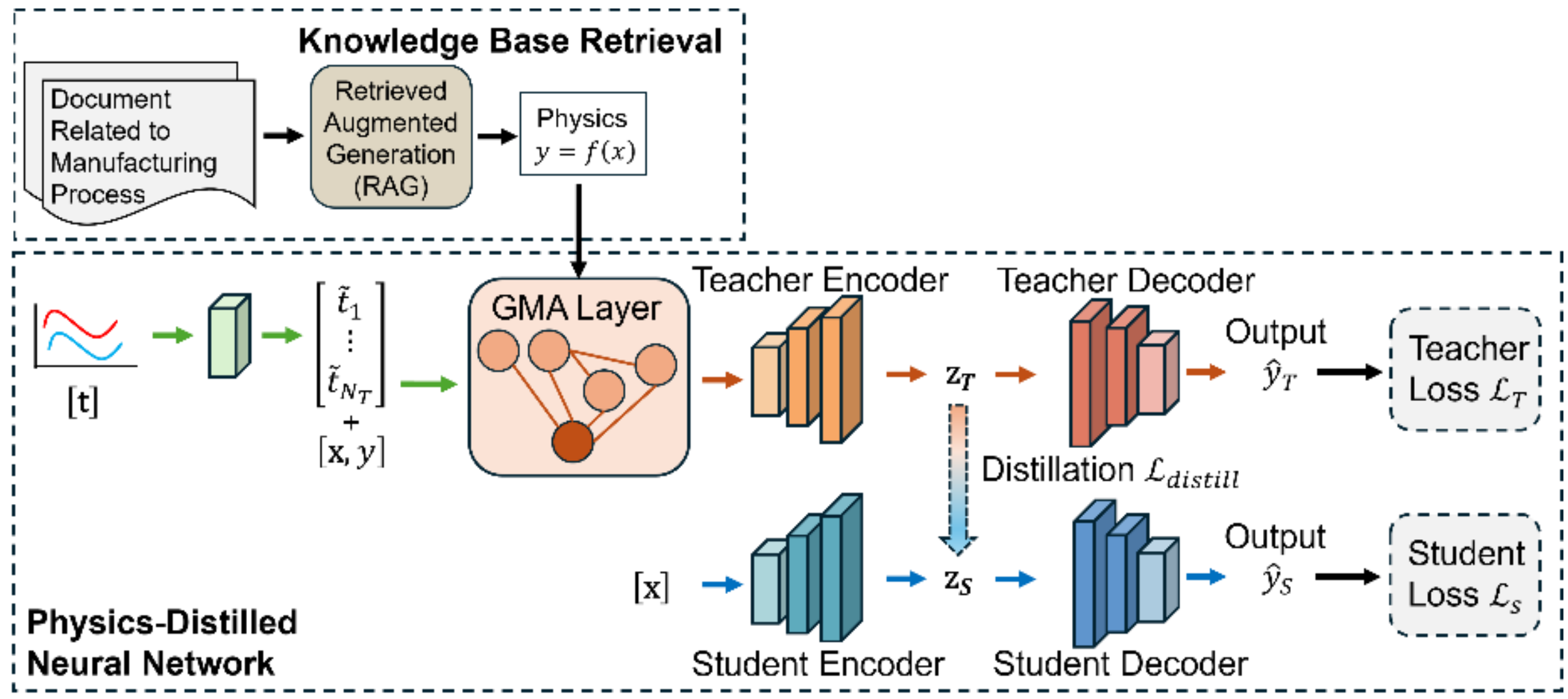

**FIGURE 1:** Overview of the proposed Physics-Distilled Neural Network framework.

### 2.1 LLM-Driven Knowledge Base Retrieval

To establish physics-grounded initial conditions for the framework, a Retrieval Augmented Generation (RAG)-based approach, based on our previous work [9], is used for obtaining parametric relationships from scientific literature on the manufacturing process of interest (Figure 2).

In the first step of this approach, research papers related to the manufacturing process are parsed and turned into text chunks, which are embedded using OpenAI's text-embedding-3-small model [48], enabling a semantic similarity search with an embedded query so that the most related chunks can be selected. Then, these reranked chunks and the original query will be sent to LLM (GPT-4o-mini [49]) to generate textual descriptions of parametric relationships. In the 'Initial Equation Generation' step, the relevant retrieved content will be synthesized by the LLM to generate initial analytical equations $f_{init}$ that capture these fundamental physical trends described in the literature. In total, 50 initial equations will be generated, fitted using 30 experimental data points, and sorted based on their performance ($R^2$ score).

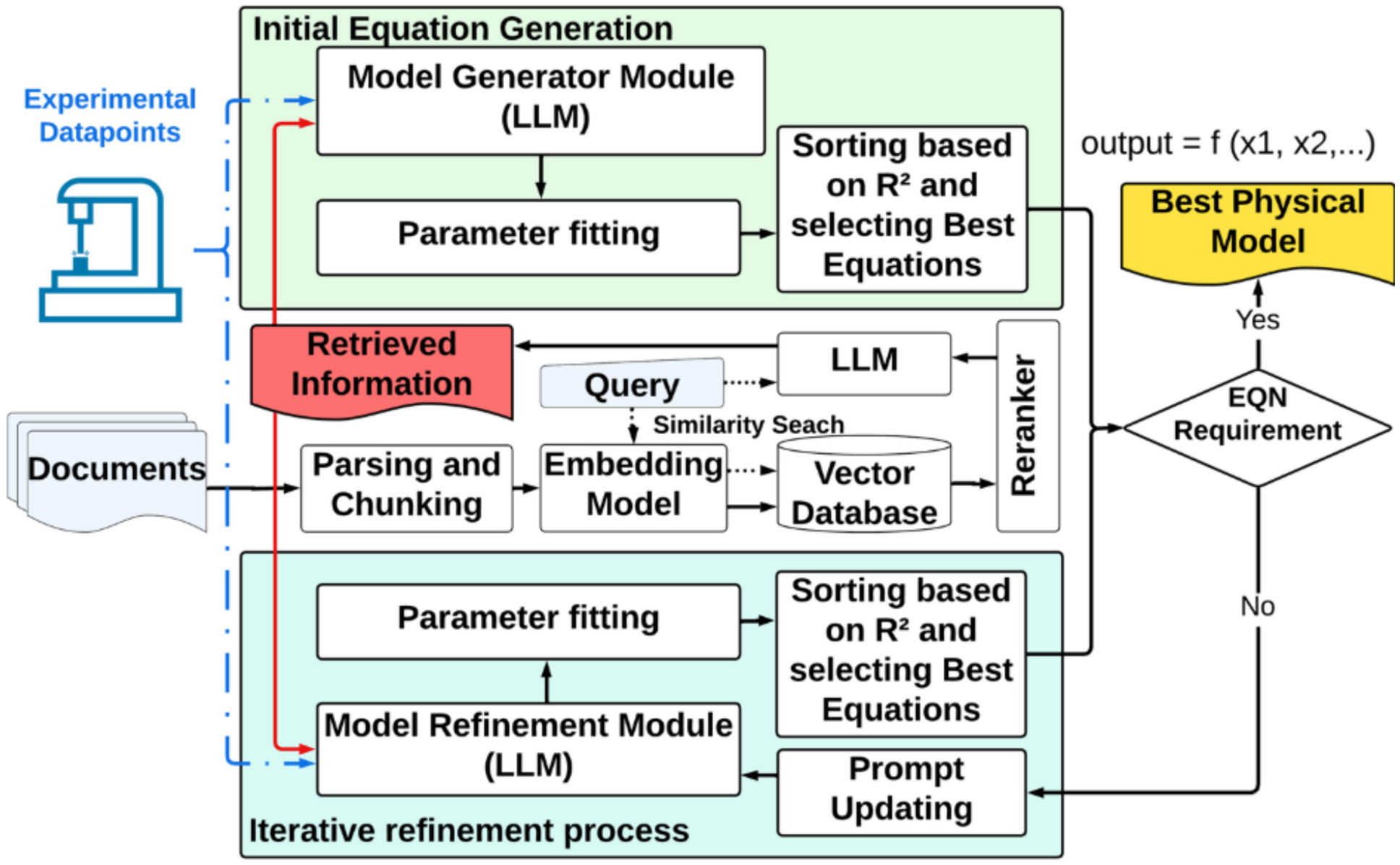

**FIGURE 2:** Overview of the LLM-driven physical model generator

If the desired accuracy has not been achieved, to enhance accuracy, an 'Iterative refinement process' is applied using a small experimental dataset, where the LLM generates 20 new candidate equations with varying complexity, evaluates them on validation data using $R^2$ scores, and iteratively refines the top-performing models through prompt-based guidance that utilizes both the retrieved physical knowledge and performance metrics ($R^2$ and MSE). This process continues until a refined equation $f_{refine}$ is obtained that satisfies the desired accuracy threshold, providing a more precise representation of the process physics.

### 2.2 Privileged Teacher Architecture

The core of the proposed framework is the *Privileged Teacher* model, which utilizes a customized self-attention architecture to integrate task-specific physical constraints directly into the neural network's decision-making process. Unlike standard models that rely solely on statistical patterns, this architecture employs a structural mask to ensure that feature interactions remain physically consistent with the governing laws of the manufacturing process.

The teacher model accepts a concatenated input vector $[\mathbf{x}, y]$, where $\mathbf{x} \in \mathbb{R}^N$ represents standard $N$ static process parameters and $y$ represents 'privileged' features (e.g., data available only during training).

The *Privileged Teacher* utilizes a specialized Graph-Masked Attention (GMA) layer where the adjacency matrix $A$ is constructed to balance empirical data-driven learning with physical constraints. We define a graph $\mathcal{G} = (\mathcal{V}, \mathcal{E})$ where $\mathcal{V}$ and $\mathcal{E}$ denote the node set and edge set, respectively. Each node $v_i \in \mathcal{V}$ represents an input feature (process parameters $\mathbf{x}$ or privileged feature $y$).

To accommodate multi-modal inputs, the framework employs a preliminary projection step for high-frequency temporal signals alongside the primary static parametric features. Specifically, if the input includes time series sequences $[\mathbf{t}] \in \mathbb{R}^{N_T \times L}$, a linear transformation is first applied to map sequences into a compact $N_T$-dimensional latent vector. Each value of this projected vector serves as the initial node embedding for the associated temporal signal, contributing to additional $N_T$ temporal nodes to the graph structure for subsequent processing.

To ensure robust convergence and a direct gradient path, the adjacency matrix $A \in \mathbb{R}^{(N+N_T+1)\times(N+N_T+1)}$ is defined by two distinct connectivity rules:

**Direct Output Enforcement:** For every input node $i \in \{x_1, \cdots, x_N\}$, we enforce a direct connection to the node $y$ to create a fail-safe in case the generated physics is partially correct. This ensures that the model always maintains a primary path to learn the direct mapping from process parameters to outcomes, regardless of the complexity of the physics:

$$A_{iy} = 1, \forall i \in \{1, \cdots, N\} \tag{1}$$

**Physics-Driven Inter-Parameter Connectivity:** The connection between input nodes $i$ and $j$ is built depending on the partial derivatives of the LLM-derived physical equation $f$. This allows the attention mechanism to prioritize physically coupled features:

$$A_{ij} = \begin{cases} 1 & \text{if } \dfrac{\partial f}{\partial x_i \partial x_j} \neq 0 \\ 0 & \text{otherwise} \end{cases} \tag{2}$$

Because every $x$ node is connected to $y$ (direct output enforcement), the model can still converge on a standard regression solution if the LLM-derived physics is incorrect. The inter-node connections act as a regularizer to force the teacher module to learn the reasons behind the outcomes, rather than just memorizing the $x \rightarrow y$ mapping.

Sequentially, a self-attention operation [50] is applied that explicitly constrains feature interactions using the physics-derived adjacency matrix $A$. Unlike standard attention, this operation prevents the model from attending to non-physical feature correlations. The model first computes the Query ($Q$), Key ($K$), and Value ($V$) vectors through learned linear transformations $W^Q$, $W^K$, and $W^V$:

$$Q/K/V = \mathbf{x}W^{Q/K/V} \tag{3}$$

These transformations allow the model to represent each feature in a way that facilitates the discovery of complex, nonlinear relationships during the attention operation. The normalized importance of each feature relationship is determined through a masked dot-product attention mechanism. More specifically, the attention scores are computed by taking the dot product of $Q$ and $K$, but the final attention weights are strictly constrained by the physical-empirical mask. For simplicity, we use the adjacency matrix $A$ as the mask in this work. The mask is applied via element-wise multiplication after the softmax operation, but before the scores are multiplied by $V$:

$$Attention(Q, K, V) = \left[softmax\left(\frac{QK^T}{\sqrt{d_k}}\right) \odot A\right] \cdot V \tag{4}$$

where $d_k$ denotes the dimension of $K$. This specific sequence ensures that any correlation assigned a weight by the data-driven softmax operation is zeroed out if it does not correspond to a valid physical or output-enforced link.

The resulting attention features are then learned and processed by a multilayer perception (MLP) based encoder to generate a latent representation $\mathbf{z}_T$. Following this, a decoder network is applied to generate the final output prediction $\hat{y}_T$. This decoder consists of multiple fully connected layers that map the high-dimensional, physics-incorporated latent features back into the target output space.

The training of the teacher model is driven by a composite loss function designed to maximize both prediction accuracy and physical consistency. The teacher loss $\mathcal{L}_T$ is defined by the summation of a prediction loss and a physics loss:

$$\mathcal{L}_T = \mathcal{L}_{pred,T} + \lambda \mathcal{L}_{phys} \tag{5}$$

The prediction loss $\mathcal{L}_{pred,T}$ utilizes mean squared error (MSE) to measure the deviation between the model output ($\hat{y}_T$) and the ground-truth ($y$). Simultaneously, the physics loss $\mathcal{L}_{phys}$ penalizes the model based on the residual error of the LLM-extracted governing equations, which acts as a soft constraint and pushes the network toward physically plausible solutions. Both loss terms are defined as follows:

$$\mathcal{L}_{pred,T} = \frac{1}{n}\sum_{i=1}^{n} \left\| \hat{y}_{T,i} - y_i \right\|^2 \tag{6}$$

$$\mathcal{L}_{phys} = \frac{1}{n}\sum_{i=1}^{n} \left\| \hat{y}_{T,i} - f(\mathbf{x}) \right\|^2 \tag{7}$$

## 2.3 Student Predictor with Knowledge Distillation

The *Student Predictor* is engineered as a lightweight, inference-ready model designed to operate in real-world manufacturing environments where the privileged ground-truth feature $y$ is unavailable. In other words, the student model relies solely on the standard process parameters $\mathbf{x}$ to generate predictions. The objective of this phase is not merely to map inputs to outputs, but to force the student to emulate the internal physical logic and feature interactions captured by the teacher's physics-guided latent manifold.

The student model utilizes a streamlined architecture consisting of an encoder and a prediction decoder. The encoder processes the primary static process parameters $\mathbf{x}$ through a series of fully connected layers, mapping them into a student latent manifold $\mathbf{z}_S$. This manifold has the same dimensionality as the teacher's latent manifold $\mathbf{z}_T$. This structural alignment is critical, as it provides the mathematical basis for direct feature-level distillation. Once the student model has learned its latent representation, a symmetric decoder network translates these features into the final prediction $\hat{y}_S$. To be more specific, the decoder of the student model has the same architecture as the decoder of the teacher model for accurate manufacturing outcome prediction.

The training of the student model is governed by the latent manifold distillation strategy, which minimizes the discrepancy between the teacher's privileged thoughts and the student's unprivileged thoughts. During the training phase, the teacher is kept in a fixed state, and its latent representation $\mathbf{z}_T$ is used as a supervisory signal. The student is optimized using a distillation loss term, $\mathcal{L}_{distill}$, which penalizes the mean squared error (MSE) between the two latent vectors:

$$\mathcal{L}_{distill} = \frac{1}{n}\sum_{i=1}^{n}\left\|\mathbf{z}_{T,i} - \mathbf{z}_{S,i}\right\|^2 \tag{8}$$

This loss forces the student to reorganize its internal feature representations to match the physics-guided manifold of the teacher. Besides the distillation loss, the prediction loss $\mathcal{L}_{pred,S}$ is also applied to ensure the model remains consistent with experimental observations:

$$\mathcal{L}_{pred,S} = \frac{1}{n}\sum_{i=1}^{n}\left\|\hat{y}_{S,i} - y_i\right\|^2 \tag{9}$$

Therefore, the total loss function for training the student model can be defined as:

$$\mathcal{L}_S = \mathcal{L}_{pred,S} + \gamma\mathcal{L}_{distill} \tag{10}$$

A primary motivation for this asymmetric knowledge distillation architecture is the practical constraint of real-time industrial deployment. While high-frequency temporal signals (e.g., acoustic emission, vibration) provide rich physical insights during model training, acquiring such data in production environments requires costly, high-bandwidth sensor arrays and significantly limits inference speed. By intentionally restricting the student predictor to utilize only static machine setpoints data that is already available from standard machine controllers, the proposed framework achieves high-speed, cost-effective inference at the edge. This approach ensures that the deployed model maintains the high predictive fidelity learned from the temporal data, without requiring the installation of auxiliary hardware on the factory floor.

## 3. EXPERIMENTAL VALIDATION

### 3.1 Experimental Manufacturing Processes

The experimental validation of the proposed framework is conducted across five diverse manufacturing domains. Initially, three mechanistically different processes, the Flow Assisted Laser-Induced Plasma Micro-Machining (FLIPMM), Masked Stereolithography (MSLA), and Turn-Assisted Deep Cold Rolling (TADCR) processes, are utilized to evaluate the framework's performance using exclusively static process setpoints (parametric inputs and outputs). To rigorously evaluate the architecture's extended multi-modal capabilities, two additional experimental processes, industrial injection molding and MaRoReS machining, are incorporated to specifically test the integration of high-frequency temporal signatures (high dimensional time series inputs) alongside static parametric input features.

The FLIPMM (Benchmark 1) process, as shown in Figure 3(a), is a subtractive micro-manufacturing technique that utilizes a laser to generate plasma within a dielectric liquid and to remove material from a workpiece while a continuous water flow washes away debris. There are four primary process parameters: pulse energy ($\mathrm{P}$, $\mu\mathrm{J}$), laser frequency ($\mathrm{F}$, $\mathrm{kHz}$), scanning speed ($\mathrm{SS}$, $\mathrm{mm/s}$), and water speed ($\mathrm{WS}$, $\mathrm{m/s}$). These controllable inputs are used to predict a critical output attribute, the heat-affected zone (HAZ, μm).

MSLA (Benchmark 2) is an additive manufacturing technique where an LCD mask selectively cures resin layers via UV-light photopolymerization to build components in a layer-by-layer manner. This process is utilized to model how parameters such as layer thickness ($\mathrm{L}$, $\mathrm{mm}$), exposure time ($\mathrm{E}$, $\mathrm{s}$), and build orientation (O, degrees) influence the ultimate tensile strength of the part (UTS, MPa). The schematic of this process is presented in Figure 3(b).

TADCR (Benchmark 3), as shown in Figure 3(c), is a deformation-based process where a rotating workpiece is compressed by a ball roller on a lathe to induce compressive residual stress and improve the fatigue life of metallic components. The framework models the influence of parameters like rolling force (R, N), ball diameter (B, mm), initial surface roughness (I, μm), and the number of rolling passes (N,-) on the output attributes of the hardness of the part (HV).

The industrial injection molding process (Benchmark 4) [51], illustrated in Figure 3(d), is a cyclical polymer manufacturing technique that utilizes high pressure to rapidly inject molten plastic, specifically Acrylonitrile Butadiene Styrene (ABS), into a precision mold cavity, where it is packed and cooled to form a solid component. There are three primary static process parameters: mold temperature (TM, °C), injection speed (S, ccm/s), and injection switchpoint (SP, ccm). These controllable inputs are used to predict a critical output attribute, the melt cushion volume (V, $cm^3$). Furthermore, high-frequency time-resolved viscometer pressure data (P, bar) is continuously recorded during each cycle to serve as privileged temporal input during model training.

The machining operation used in the MaRoReS (Benchmark 5) dataset [52] is a subtractive turning process that utilizes a lathe to continuously shear material from 42CrMo4+QT alloy steel workpieces. In this specific domain, the static parametric features guiding the operation include the feed rate (FR, mm/rev), cutting speed (S, m/min), depth of cut (D, mm), and insert radius (R, mm). The dynamic features, which serve as the privileged temporal information during model training, consist of high-frequency measurements of cutting forces (F, N), tool holder accelerations (A, $mm/s^2$), sound impact (SI, dB), and current consumed by the lathe (C, A). The final target for prediction is the radial residual surface stresses (RS, MPa) induced within the machined workpiece.

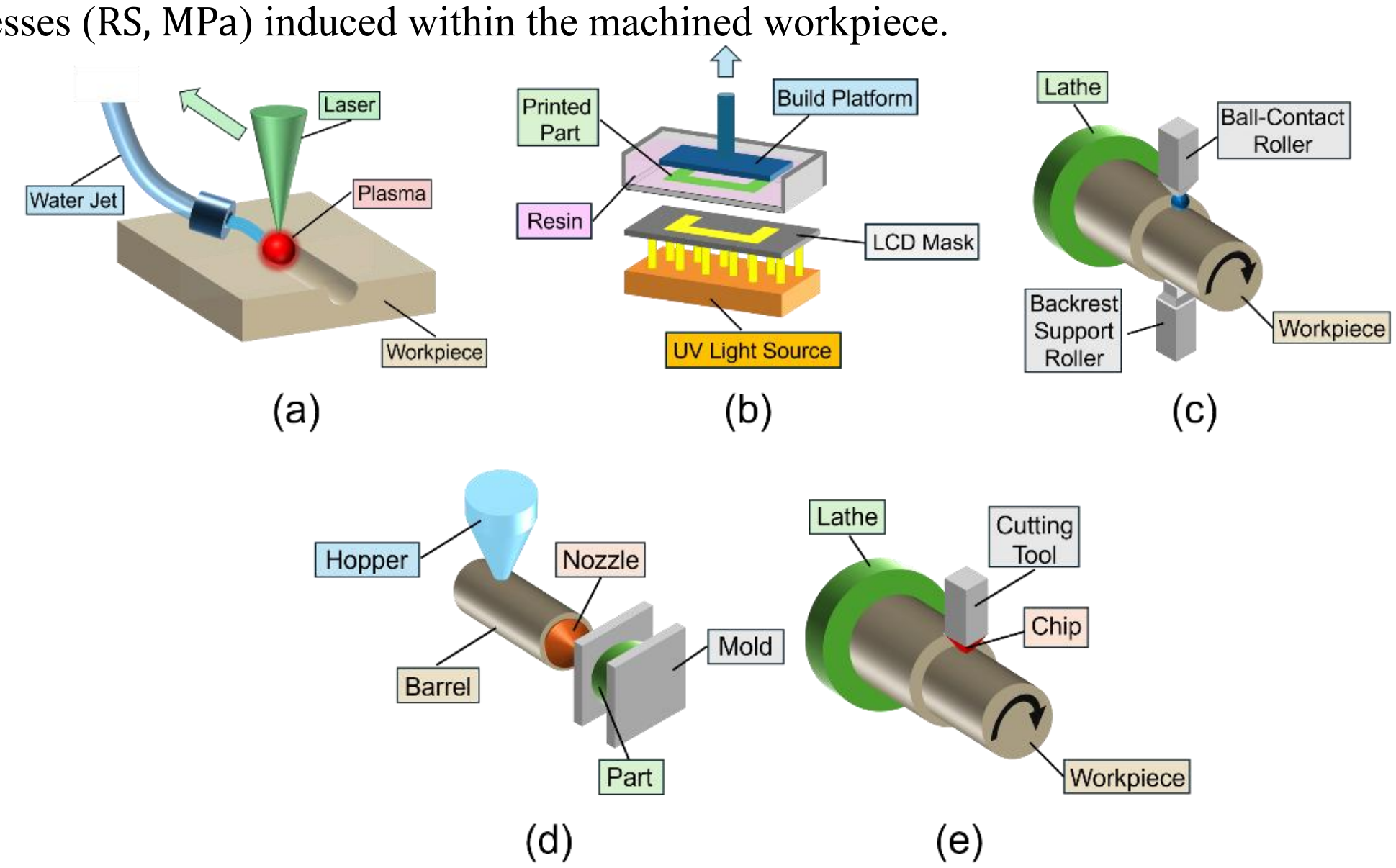

**FIGURE 3:** The schematic of 5 benchmarks

### 3.2 Dataset Preparation

The knowledge extraction step of this work utilized a comprehensive repository consisting of 17 publications for FLIPMM (Benchmark 1), 41 for MSLA (Benchmark 2), 10 for TADCR (Benchmark 3), 10 for injection molding (Benchmark 4), and 10 for MaRoReS (Benchmark 5). These sources were obtained from Google Scholar by identifying process-related keywords and screening abstracts for general relevance.

The five benchmarks span two distinct data regimes. Benchmarks 1-3 use only static parametric inputs, with ground-truth labels generated from validated Response Surface Methodology (RSM) equations established in prior works [53] [54] [55] [56]. These are evaluated on a factorial grid in the input parameter space. Benchmarks 4 and 5 use real experimental data that additionally includes high-dimensional time-series measurements available only during training as privileged information. All input features and target outputs are standardized using StandardScaler to stabilize convergence and mitigate scale disparities across the diverse physical measurements.

For Benchmarks 1-3, we intentionally keep the training set small to emulate the severe data scarcity characteristic of industrial settings. A hierarchical extrapolation split is applied to the synthetic factorial grid. The candidate training/validation pool is defined as the set of grid points whose values lie within the inner 75% range of every input variable. All remaining points, those outside the inner 75% in at least one dimension, form the test set. From the candidate training pool, 30 training samples are randomly selected. This random subset sampling is repeated five times with different data-sampling seeds, while the test set is held fixed across runs. This split rigorously evaluates extrapolation, a known limitation of data-driven manufacturing models [57]. It also yields a distribution of accuracy performance per model to enable statistical comparison. For Benchmarks 4 and 5, the experimental datasets contain only 68 instances each. A 5-fold cross-validation scheme is therefore used. Table 1 summarizes the per-benchmark configuration.

**TABLE 1:** Dataset configuration for each benchmark process

| Benchmark | Input type | Total instances | Training | Test | Split strategy |
|---|---|---|---|---|---|
| 1 | Static parameters | 256 ($4^4$ grid) | 30 | 175 | Hierarchical extrapolation |
| 2 | Static parameters | 216 ($6^3$ grid) | 30 | 152 | Hierarchical extrapolation |
| 3 | Static parameters | 256 ($4^4$ grid) | 30 | 175 | Hierarchical extrapolation |
| 4 | Static+time-series | 68 | ~54/fold | ~13/fold | 5-fold CV |
| 5 | Static+time-series | 68 | ~54/fold | ~13/fold | 5-fold CV |

### 3.3 Model Training, Hyperparameter Configuration, and Evaluation Metrics

The training of the proposed framework was executed using a dual-network optimization strategy to balance empirical data fitting with physical consistency. Optimization was performed using the separate AdamW optimizers for both the teacher model and the student model with a learning rate of 0.001 and 0.01, respectively. The training was conducted for 1000 epochs using a batch size of 16 to ensure sufficient iterations for the student model to distill the latent representations from the teacher. The weight of the physical loss term in Eq. 5 is set to 0.1 ($\lambda = 0.1$), while a coefficient of 2 was assigned to the distillation loss in Eq. 10 ($\gamma = 2$).

To quantitatively evaluate the prediction performance of the proposed framework, we utilize the Coefficient of Determination $R^2$ and Root Mean Squared Error ($RMSE$). These metrics provide a dual perspective on the model's accuracy. The $R^2$ score indicates the proportion of variance captured by the model, while $RMSE$ provides a direct measure of the average deviation from experimental values. A higher value of $R^2$ score indicates better prediction, while a lower value of RMSE means better predictive performance. The Coefficient of Determination and Root Mean Squared Error is defined as follows:

$$R^2 = 1 - \frac{\sum_{i=1}^{n}\left(y_i - \hat{y}_{S,i}\right)^2}{\sum_{i=1}^{n}(y_i - \bar{y})^2} \tag{11}$$

$$RMSE = \sqrt{\frac{1}{n}\sum_{i=1}^{n}\left(y_i - \hat{y}_{S,i}\right)^2} \tag{12}$$

where $y_i$ represents the experimental ground truth, $\hat{y}_{S,i}$ is the predicted value from the student predictor, $\bar{y}$ is the mean of the experimental data, and $n$ is the total number of test samples.

In addition to prediction accuracy, the ability of the model for real-time manufacturing monitoring and in-situ process control depends heavily on its computational efficiency. To quantify this, the inference speed of the student predictor is evaluated. Since the inference time can vary slightly across individual execution cycles due to system overhead, the inference time is measured over 10 independent runs, and the average value is reported. The calculation frequency ($v$) is then defined as the inverse of the average inference time:

$$v = \frac{1}{\bar{t}_{inference}} \tag{13}$$

### 3.4 Physics Prior Conditions

To demonstrate the framework's robustness, we evaluated the framework using two levels of physical knowledge. Specifically, we utilized two distinct equations extracted from the literature via the RAG-based framework described in the previous work [9]: an initial equation $f_{init}$ representing raw, unrefined literature trends, and a refined equation $f_{refine}$ that has undergone iterative optimization using experimental data to achieve a better fit to the training data.

Table 2 lists the ten analytical priors extracted by the RAG-based framework [9] across five benchmarks. For each benchmark, the initial equation $f_{init}$ represents the raw, unrefined trend distilled directly from the literature, and the refined equation $f_{refine}$ is obtained through iterative optimization. The standalone prediction accuracy of each equation, evaluated on the test set, is reported alongside the expression. Notably, given that 5-fold cross-validation is applied on Benchmark 4 and 5, Equations 20-23 show the equations generated for the representative fold (the fold whose test set $R^2$ is closest to the 5-fold mean). The $R^2$ values reported are the means across all five folds. For Benchmarks 4 and 5, each temporal sensor channel (time-series sequence) is reduced to its average scalar value (denoted as $\bar{\cdot}$ in the equations).

**TABLE 2:** Analytical physical priors extracted by the RAG-based framework

| Eq. | Benchmark | Variant | Expression | $R^2$ |
|---|---|---|---|---|
| (14) | 1 | $f_{init}$ | $0.2941 \cdot P + 5.6005 \cdot \frac{1}{SS} + 166.305 \cdot \frac{1}{WS} + 0.0414 \cdot F^2$ | 0.689 |
| (15) | 1 | $f_{refine}$ | $0.0013 \cdot P^2 + 5.5453 \cdot \sqrt{\frac{1}{SS}} + 235.051 \cdot \frac{1}{WS} - 0.1852 \cdot F^{1.5} + 0.131 \cdot P \cdot F - 0.6376 \cdot SS \cdot \sqrt{\frac{1}{WS}} + 0.0011 \cdot P \cdot F^2 - 0.0002 \cdot P^{1.2} \cdot F \cdot SS - 0.6521 \cdot F^{0.5} \cdot P^{0.8}$ | 0.928 |
| (16) | 2 | $f_{init}$ | $50.2946 + 8.1617 \cdot \ln E - 155.9657 \cdot L - 3.4706 \cdot \sin^2(\text{radians}(O))$ | 0.622 |
| (17) | 2 | $f_{refine}$ | $43.6926 - 39.6988 \cdot \sqrt{L} + 25.9001 \cdot \ln(E+1) - 10.8882 \cdot \sqrt{E} - 1.5444 \cdot \sin(\text{radians}(O)) \cdot \left(1 - \frac{(O-60)^2}{900}\right) + 0.0006 \cdot (O-45)^2 - 592.7573 \cdot \frac{L}{10}$ | 0.708 |
| (18) | 3 | $f_{init}$ | $4.9965 \cdot B + 0.0671 \cdot R - 2.8784 \cdot I + 7.5271 \cdot N + 203.8603$ | 0.936 |
| (19) | 3 | $f_{refine}$ | $229.7337 + 0.0972 \cdot B^2 + 0.0000055 \cdot R^2 - 0.2488 \cdot I^2 + 7.1043 \cdot N + 0.0081 \cdot B \cdot R + 0.0081 \cdot B \cdot N - 0.0081 \cdot I \cdot N$ | 0.957 |
| (20) | 4 | $f_{init}$ | $-0.0021 - 0.1231 \cdot TM + 0.0262 \cdot S - 0.0277 \cdot SP + 0.1663 \cdot \bar{P}$ | 0.927 |
| (21) | 4 | $f_{refine}$ | $-4.0867 + 0.0028 \cdot TM \cdot P - 9.7631 \cdot \frac{SP}{S} + 0.000047 \cdot \bar{P}$ | 0.942 |
| (22) | 5 | $f_{init}$ | $-215.8493 - 240.9857 \cdot R - 282.5206 \cdot D + 1.0819 \cdot S + 3729.1541 \cdot FR + 104.3688 \cdot \bar{F} + 66.3097 \cdot \bar{A} - 3.4503 \cdot \overline{SI} + 18.6272 \cdot \bar{C}$ | 0.393 |
| (23) | 5 | $f_{refine}$ | $197.8487 - 22679.1818 \cdot FR^2 - 2032.7464 \cdot \sqrt{D} - 318.8555 \cdot R + 0.8640 \cdot S + 119.4057 \cdot \bar{F} + 14983.5418 \cdot FR \cdot D^{0.2} + 92.6953 \cdot \bar{A} - 3.9145 \cdot \bar{C} \cdot FR + 0.3772 \cdot \frac{\overline{SI}}{R}$ | 0.515 |

As shown in the table, initial equations typically capture only first-order trends and exhibit substantial fitting error (notably $R^2 < 0.7$ for Benchmarks 1, 2, and 5). After RAG-driven refinement, every benchmark shows improvement. The refined priors are still far from individually accurate on the experimental Benchmarks 5, which motivates the distillation framework. For Benchmarks 3 and 4, even initial equations already achieve high accuracy with values of $R^2$ over 0.9. The corresponding scatter plots (Figure 4) also show that the equations have limitations in qualitatively different ways across benchmarks, e.g., systematic bias for FLIPMM, discretized predictions for injection molding, and physically implausible outputs for a substantial fraction of MaRoReS cases.

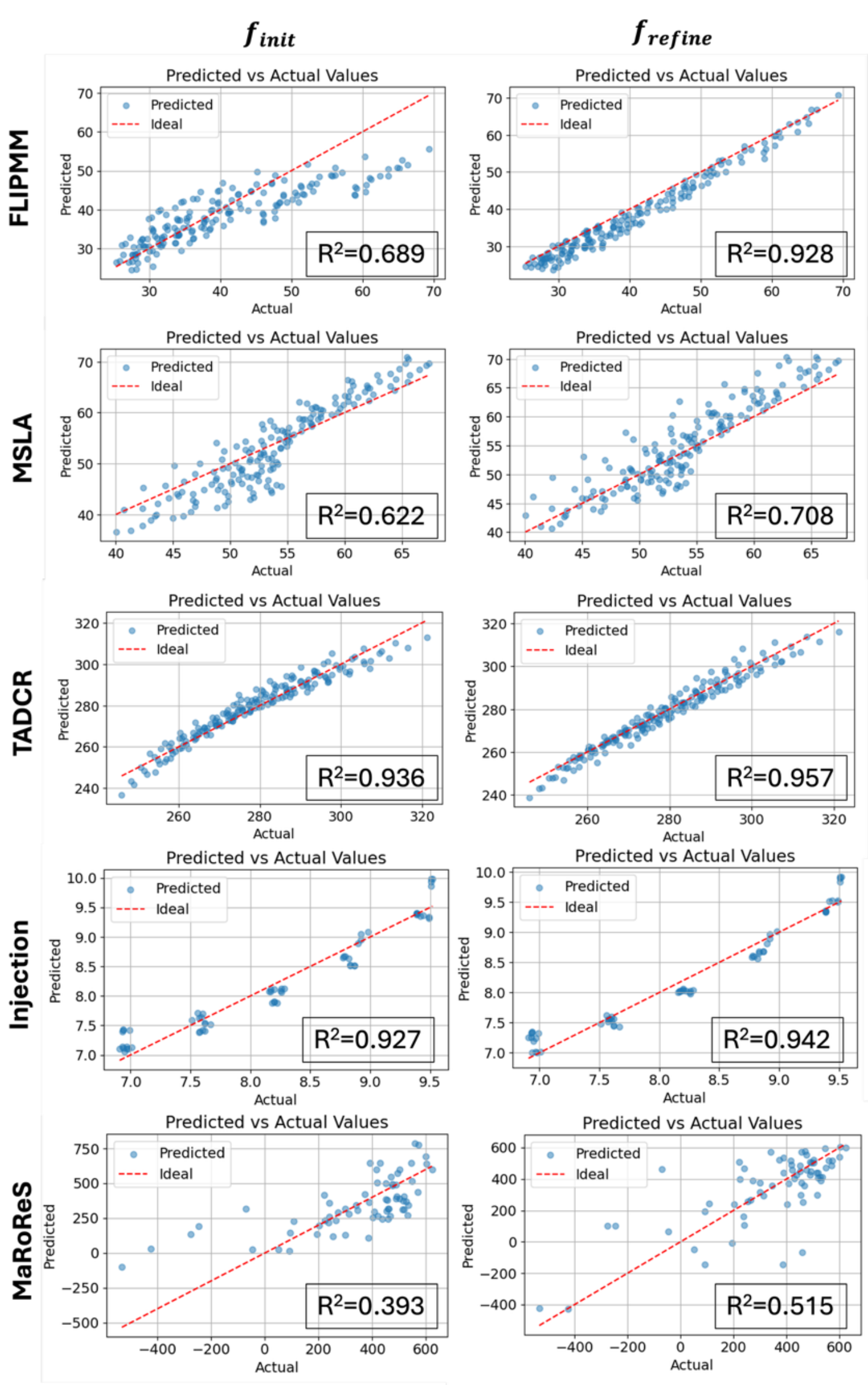

**FIGURE 4:** Standalone prediction accuracy of the analytical physical equations for all testing cases

### 3.5 Result Analysis and Discussion

The mean quantitative evaluation results are summarized in Table 3, which provides a detailed breakdown of the $R^2$ and $RMSE$ scores. The values confirm that while the initial physics model already provides a reliable prediction performance, leveraging refined physical priors leads to a slight increase in $R^2$ and a corresponding reduction in absolute error. These metrics validate the efficiency of the knowledge distillation approach by showing that the student model effectively learns the optimized physical representations to achieve a level of accuracy that bridges the gap between general analytical laws and the specific experimental conditions of the testing manufacturing processes.

**TABLE 3:** Performance summary of the proposed framework

| Evaluation | $f_{init}/f_{refine}$ | | | | |
|---|---|---|---|---|---|
| Process | Benchmark 1 | Benchmark 2 | Benchmark 3 | Benchmark 4 | Benchmark 5 |
| $R^2$ ↑ | 0.949/0.957 | 0.824/0.826 | 0.957/0.964 | 0.963/0.989 | 0.612/0.725 |
| $RMSE$ ↓ | 2.410/2.204 | 2.516/2.505 | 3.240/3.109 | 0.169/0.099 | 0.498/0.347 |

To rigorously validate the proposed framework according to the requirements of high-precision manufacturing modeling, a comprehensive benchmarking study was conducted against three distinct categories of machine learning: traditional regression, traditional neural networks, and physics-informed neural networks. The category of traditional regression includes Random Forest (RF) [58] and XGBoost [59]. The Multi-Layer Perceptron (MLP) [60] is used as a classical traditional neural network, which shares the exact same model architecture as the *Student Predictor* of the proposed framework for a fair comparison. A traditional knowledge distillation (KD) [61] network is further included as one baseline. This network employs an identical teacher-student configuration but with the GMA layer and physical loss terms removed. Finally, a Physics-Guided MLP (PG-MLP) [62] is evaluated, which incorporates the same physical loss term as our student model but lacks the GMA layer. These models were trained using the same sparse training dataset to ensure a fair comparison under identical data constraints. Beyond standard benchmarking, these models can also function as an ablation study to isolate the performance contribution of each architectural part. Specifically, comparing the proposed framework to the KD network verifies the effectiveness of the GMA layer in providing structural guidance. Comparing with the traditional PG-MLP network demonstrates the impact of removing privileged information during model training and knowledge distillation. Since inference speed is determined by the model architecture and input dimensionality, and these are essentially constant within each data regime, the speed comparison is reported only twice: once for the static parametric input-only benchmarks using FLIPMM (Benchmark 1) as the representative case, and once for the privileged-information benchmarks using injection molding (Benchmark 4).

**TABLE 4:** Comprehensive comparative benchmarking for FLIPMM (Benchmark 1)

| Category | Model | Mean Accuracy ($R^2/RMSE$) | | Speed (HZ) |
|---|---|---|---|---|
| | | $f_{init}$ | $f_{refine}$ | |
| Physics-only | Physics | 0.689/5.939 | 0.928/2.867 | 4332 |
| Traditional Regression | RF | 0.342/8.645 | | 1726 |
| | XGBoost | 0.453/7.886 | | 969 |
| Traditional Neural Network | MLP | 0.896/3.432 | | 6494 |
| | KD | 0.907/3.248 | | 6572 |
| Physics-Incorporated | PG-MLP | 0.854/4.073 | 0.868/3.872 | 6511 |
| | Proposed | 0.949/2.410 | 0.957/2.204 | 6537 |

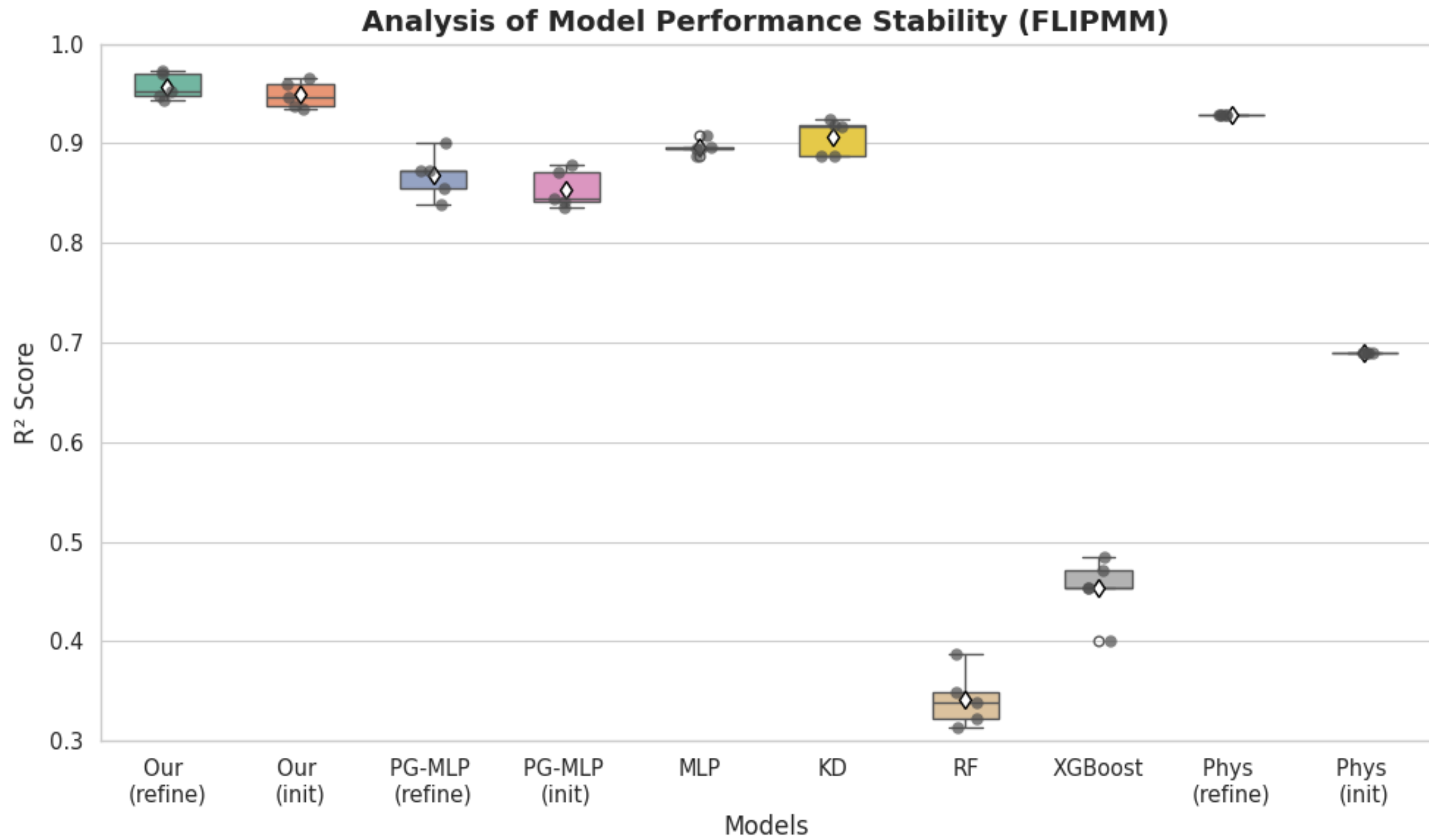

**FIGURE 5:** Performance comparison of prediction models for FLIPMM (Benchmark 1)

Table 4 and Figure 5 show the model comparison for Benchmark 1. The proposed framework achieves the highest mean $R^2$ (0.957 refined, 0.949 initial) while also exhibiting the tightest interquartile range among trainable models, indicating both accuracy and stability against variations in the training subset. Notably, the refined physics equation alone reaches $R^2 = 0.928$, exceeding all data-only neural baselines (MLP, KD, PG-MLP). This confirms that the LLM-distilled analytical prior already captures the dominant trends, and the proposed framework's additional gain comes from compensating for the equation's residual error rather than learning the process from scratch. The two traditional regression models, RF and XGBoost, fall sharply below the rest with $R^2 < 0.5$, reflecting their difficulty in extrapolating beyond the inner-75% training region. It is also worth noting that the physics-only baselines appear as single horizontal markers for Benchmark 1-3 because these are analytical equations evaluated on a fixed extrapolation test set and with no training involved. The result is deterministic across runs.

**TABLE 5:** Comprehensive comparative benchmarking for MSLA (Benchmark 2)

| Category | Model | Mean Accuracy ($R^2/RMSE$) | |
|---|---|---|---|
| | | $f_{init}$ | $f_{refine}$ |
| Physics-only | Physics | 0.622/3.697 | 0.708/3.251 |
| Traditional Regression | RF | 0.762/2.936 | |
| | XGBoost | 0.713/3.076 | |
| Traditional Neural Network | MLP | 0.781/2.834 | |
| | KD | 0.804/2.693 | |
| Physics-Incorporated | PG-MLP | 0.793/2.767 | 0.807/2.679 |
| | Proposed | 0.824/2.516 | 0.826/2.505 |

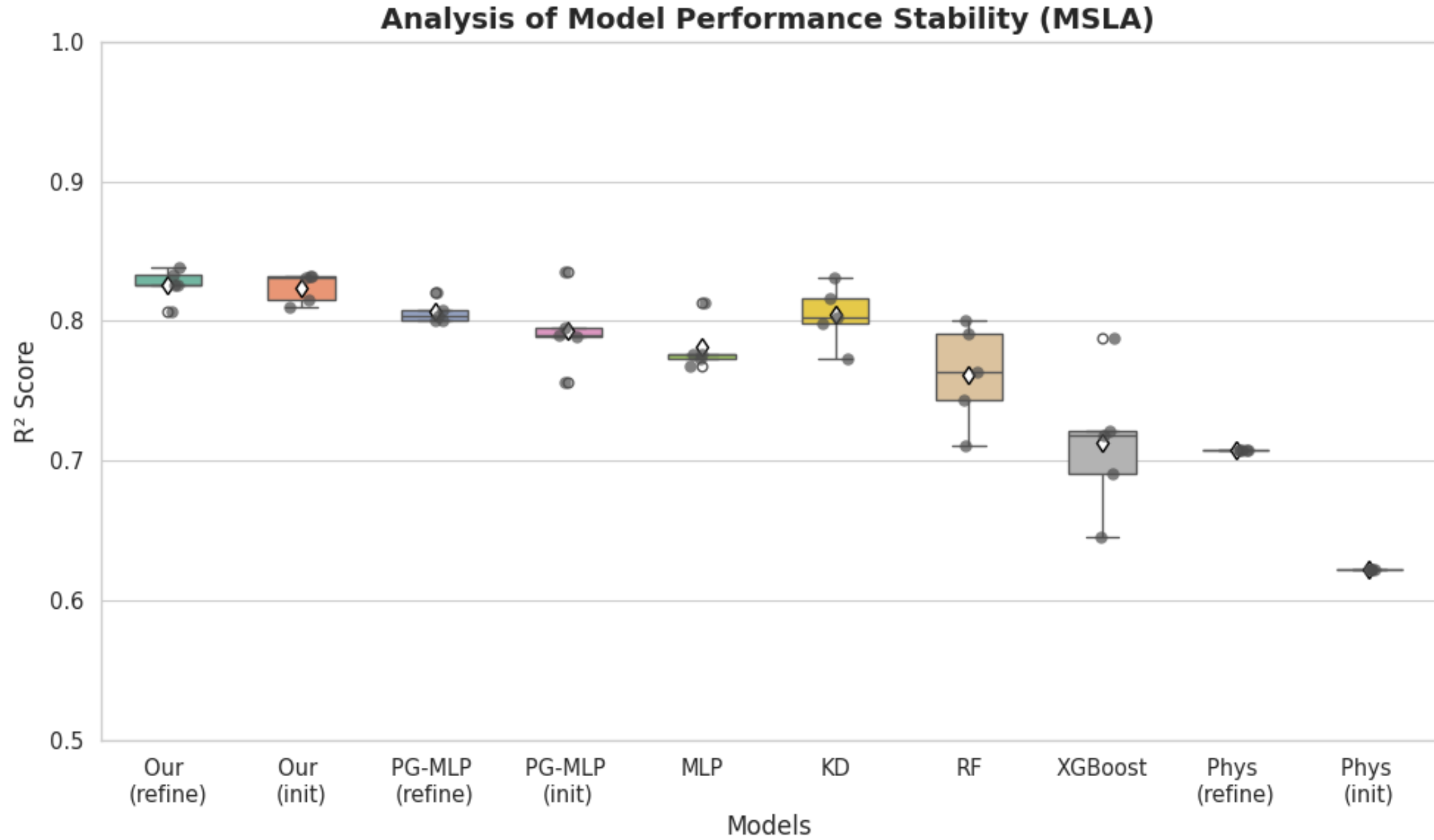

**FIGURE 6:** Performance comparison of prediction models for MSLA (Benchmark 2)

The comparative performance for the Benchmark 2 is detailed in Table 5 and Figure 6. The physics-only models perform poorly in this case ($R^2$ of 0.622 and 0.708 for the initial and refined equations). The proposed framework nonetheless leads with $R^2 = 0.826$ (refined), with the interquartile range narrower than that of MLP and KD, supporting the claim that the framework extracts useful guidance even from a weak analytical prior. The narrow performance gap between PG-MLP and the proposed model further indicates that physical guidance through masked attention is more effective than simple loss-based constraints when the underlying physics equation is itself imperfect.

**TABLE 6:** Comprehensive comparative benchmarking for TADCR (Benchmark 3)

| Category | Model | Mean Accuracy ($R^2/RMSE$) | |
|---|---|---|---|
| | | $f_{init}$ | $f_{refine}$ |
| Physics-only | Physics | 0.936/3.987 | 0.957/3.262 |
| Traditional Regression | RF | 0.620/7.498 | |
| | XGBoost | 0.794/6.535 | |
| Traditional Neural Network | MLP | 0.944/3.798 | |
| | KD | 0.951/3.331 | |
| Physics-Incorporated | PG-MLP | 0.946/3.514 | 0.949/3.427 |
| | Proposed | 0.957/3.240 | 0.964/3.109 |

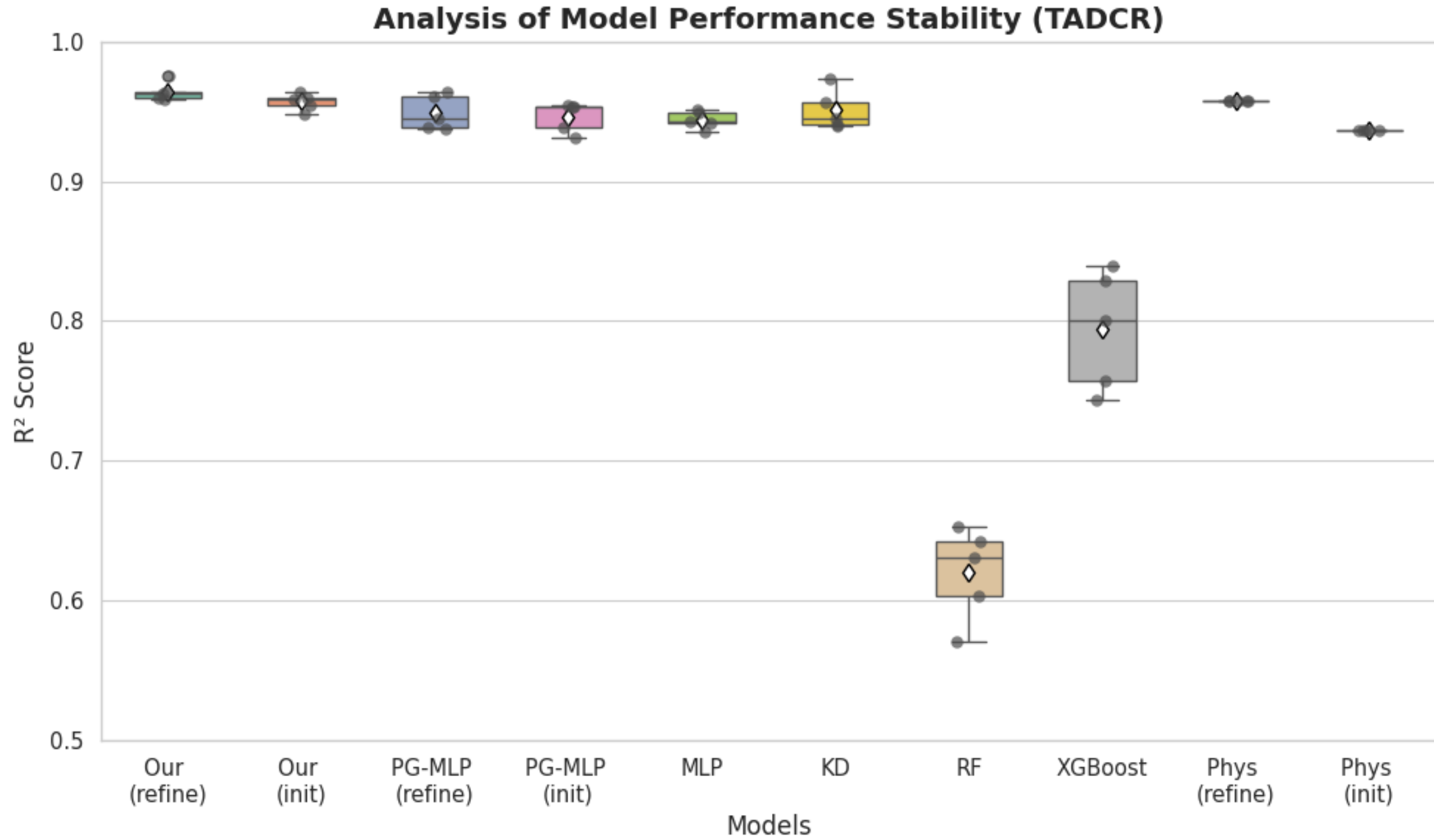

**FIGURE 7:** Performance comparison of prediction models for TADCR (Benchmark 3)

The Benchmark 3 results are presented in Table 6 and Figure 7. This benchmark represents the opposite extreme of Benchmark 2 where the refined analytical equation is already highly accurate ($R^2 = 0.957$) due to the maturity of the literature on which it is based, raising the effective performance ceiling for all methods. Most neural networks, including MLP, KD, PG-MLP, and the proposed model, converge to a high level of $R^2$ with tight interquartile ranges, while RF and XGBoost lag substantially. The proposed framework remains the top performer, confirming that even when the analytical prior is strong, the architecture extracts incremental but consistent improvement. This result also indicates the framework does not degrade in regimes where physics alone suffices.

**TABLE 7:** Comprehensive comparative benchmarking for injection molding (Benchmark 4)

| Category | Model | Mean Accuracy ($R^2/RMSE$) | | Speed (HZ) |
|---|---|---|---|---|
| | | $f_{init}$ | $f_{refine}$ | |
| Physics-only | Physics | 0.927/0.229 | 0.942/0.200 | 3987 |
| Traditional Regression | RF | 0.926/0.149 | | 1180 |
| | XGBoost | 0.683/0.470 | | 1079 |
| Traditional Neural Network | MLP | 0.969/0.141 | | 6138 |
| | KD | 0.971/0.133 | | 6191 |
| Physics-Incorporated | PG-MLP | 0.929/0.288 | 0.970/0.137 | 6207 |
| | Proposed | 0.963/0.169 | 0.989/0.099 | 6198 |

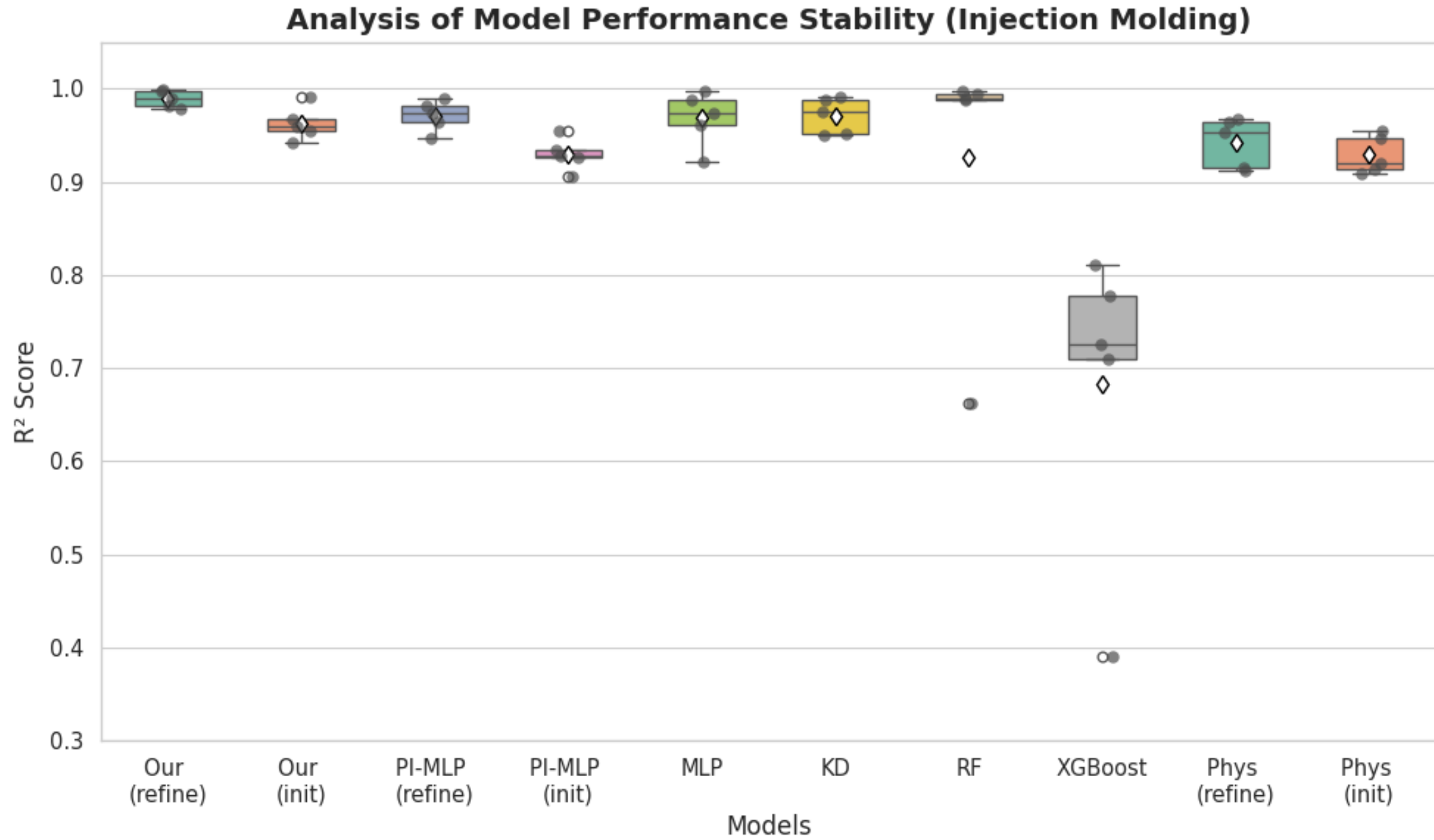

**FIGURE 8:** Performance comparison of prediction models for injection molding (Benchmark 4)

The predictive performance for the melt cushion volume within the injection molding dataset (Benchmark 4) highlights the advantages of the proposed framework, as detailed in the quantitative metrics table (Table 7) and the cross-validation box plot (Figure 8). Standard data-driven baselines, including the conventional MLP and XGBoost models, exhibit lower average $R^2$ scores alongside higher $RMSE$. This reflects their inherent difficulty in extracting generalized patterns from the strictly limited pool of physical experimental samples. Furthermore, the box plot, which illustrates the $k$-fold validation $R^2$ scores, reveals variance for these prediction methods. The plot indicates a high susceptibility of other models to overfitting and instability depending on the specific data partition. Conversely, our proposed model achieves the highest overall predictive accuracy and the lowest error margins while demonstrating exceptional statistical stability with a tightly constrained interquartile range across all validation folds. This confirms its robust generalization capabilities and resistance to sampling bias across the complex manufacturing parameter space.

**TABLE 8:** Comprehensive comparative benchmarking for MaRoReS (Benchmark 5)

| Category | Model | Mean Accuracy ($R^2/RMSE$) | |
|---|---|---|---|
| | | $f_{init}$ | $f_{refine}$ |
| Physics-only | Physics | 0.393/1.257 | 0.515/0.693 |
| Traditional Regression | RF | 0.605/0.515 | |
| | XGBoost | 0.396/0.675 | |
| Traditional Neural Network | MLP | 0.425/0.653 | |
| | KD | 0.515/0.579 | |
| Physics-Incorporated | PG-MLP | 0.478/0.611 | 0.579/0.534 |
| | Proposed | 0.612/0.498 | 0.725/0.347 |

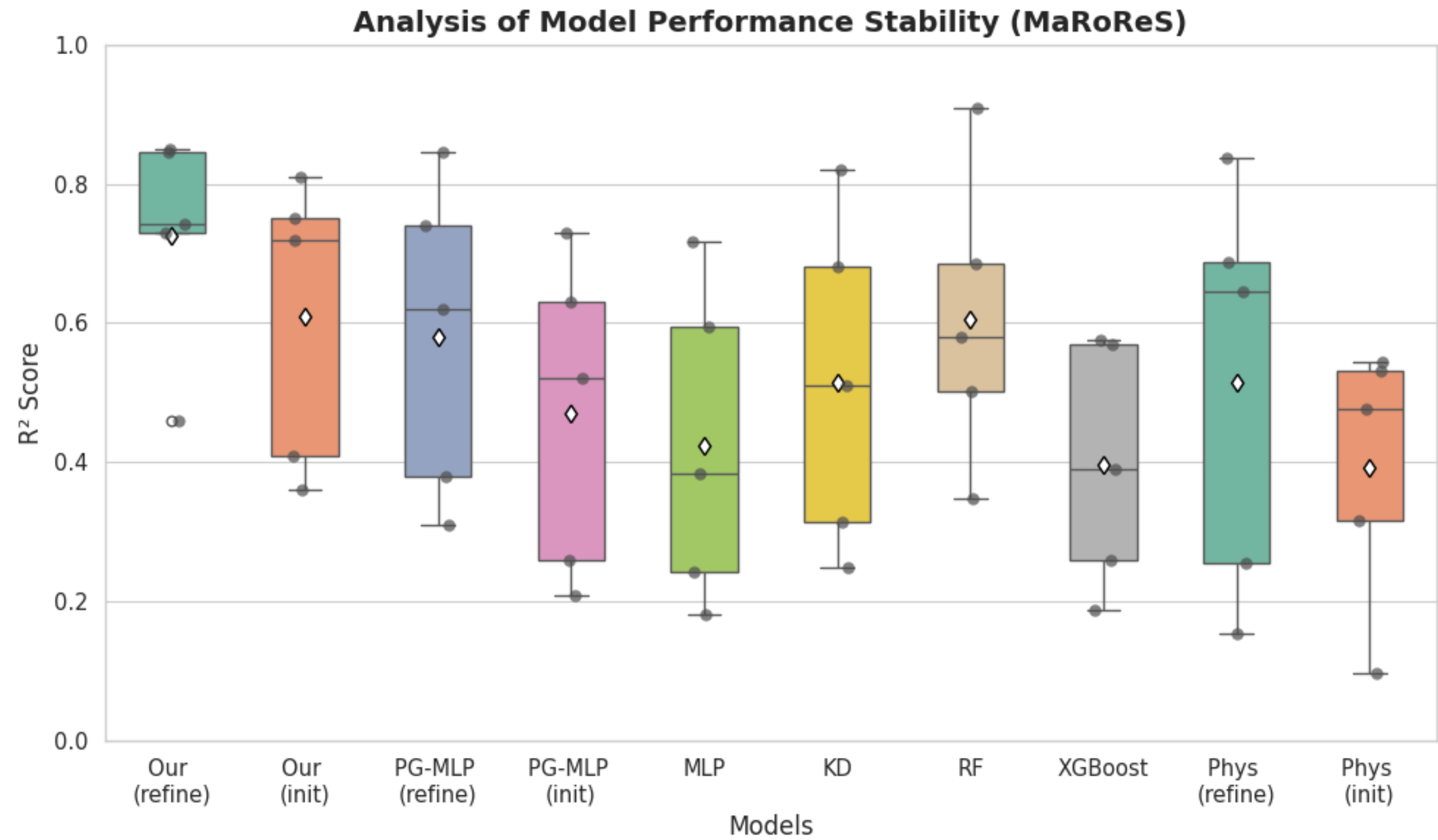

**FIGURE 9:** Performance comparison of prediction models for MaRoReS (Benchmark 5)

The evaluation of radial residual stress prediction within the MaRoReS dataset (Benchmark 5) clearly validates the efficiency of combining physical laws with dynamic sensor data, as evidenced by the performance metrics (Table 8) and cross-validation distributions (Figure 9). Purely data-driven architectures face severe limitations when attempting to model this highly non-linear tuning process with restricted sample sizes. For example, the standard MLP and XGBoost baselines yield average $R^2$ scores of just 0.425 and 0.396, respectively. The associated box plot further exposes the limitation of these conventional methods, showing that the MLP's accuracy collapses to an $R^2$ of 0.182 on the most challenging data partition. In contrast, the proposed architecture achieves a dominant mean $R^2$ of 0.725. The significantly compressed interquartile range in the box plot proves that the physical constraints successfully anchor the model to ensure reliable and stable generalization even during the most difficult cross-validation splits.

## 4. CONCLUSION

This paper introduces a Physics-Distilled Teacher-Student framework for accurate, lightweight, and robust process-property in data-scarce manufacturing settings. The framework integrates three architectural components: (1) an LLM-driven retrieval-augmented generation pipeline that automatically extracts and refines analytical priors from the process-specific literature, removing the need for manual derivation of physical loss terms; (2) a privileged teacher network built on a Graph-Masked Attention layer, whose attention mask is structurally derived from the symbolic prior so that the learned representation respects the variable-interaction topology implied by physics; and (3) a lightweight student predictor trained by feature-level distillation from the teacher, retaining the teacher's physics-aligned inductive bias while remaining deployable at high inference rates.

The novelty of this work lies in three contributions that distinguish it from existing physics-incorporated and knowledge-distilled approaches for manufacturing. First, prior physics-informed neural networks for manufacturing rely on human-derived physical loss terms, limiting their scalability across processes. In this work, the analytical prior is automatically extracted by an LLM-RAG pipeline. Second, existing knowledge-distillation studies in manufacturing address fault-diagnosis classification under abundant data, whereas this work applies distillation to small-data regression with privileged temporal information available only during training, which is a setting more representative of industrial process modeling. Third, the framework integrates the analytical prior at the architectural level (via the attention mask) rather than only at the loss level. It provides a structural inductive bias that proves more effective than loss-based physics constraints in our comparisons.

Several directions remain open for future work. First, the predicted outputs in this study are scalar quantities. Extending the framework to spatial-field outputs, such as full residual-stress fields or distortion maps, would broaden its applicability to design-for-manufacturing tasks beyond point-wise quality prediction. In addition, the framework currently produces point predictions. Integrating calibrated uncertainty quantification would enable risk-aware deployment in process-control loops where the cost of overconfidence is high.

## ACKNOWLEDGEMENTS

The authors acknowledge financial support from NSF CMMI-2414398, CMMI-2414397 and CMMI-2336448; U.S. Department of Energy's Office of Energy Efficiency and Renewable Energy (EERE) under the Advanced Manufacturing Office Award Number DEEE0011029.

## Appendix

### 1. Detailed Neural Network Architecture

To ensure reproducibility, the detailed network architectures and layer dimensions for both the Privileged Teacher and the Student Predictor across all five manufacturing processes are provided below. Because some datasets are strictly limited in size, the hidden layer dimensions are intentionally kept lightweight to prevent severe overfitting. The latent manifold dimension ($z$) is fixed at 16 across all models to ensure a stable distillation target. All five benchmarks use the same architectural template. Only the input dimensionality varies across processes.

**TABLE 9:** Layer-by-layer specification of the Privileged Teacher

| Module | Layer | Input→Output dim | Notes |
|---|---|---|---|
| Temporal projector | Linear+LeakyReLU | $1000 \rightarrow 1$ | Applied per time-series channel; Benchmark 4 & 5 only |
| GMA layer | Masked self-attention | $(N + N_T + 1) \rightarrow (N + N_T + 1)$ | Adjacency from physics; heads=4 |
| Encoder | Linear+LeakyReLU+Linear | $(N + N_T + 1) \rightarrow 8 \rightarrow 16$ | |
| Decoder | Linear+LeakyReLU+Linear | $16 \rightarrow 8 \rightarrow 1$ | |

**TABLE 10:** Layer-by-layer specification of the Student Predictor

| Module | Layer | Input→Output dim | Notes |
|---|---|---|---|
| Encoder | Linear+LeakyReLU+Linear | $N \rightarrow 8 \rightarrow 16$ | Same latent dimension as Teacher |
| Decoder | Linear+LeakyReLU+Linear | $16 \rightarrow 8 \rightarrow 1$ | Identical to Teacher decoder |

### 2. Knowledge Retrieval Query and Model Generation Prompt

In this section a summarized Query Form used in RAG model to extract parametric knowledge from the literature and a summarized Prompt Form used to translate that knowledge into analytical Python models with iterative refinement has been included. For complete form, please refer to [9].

The query is submitted to the RAG system to semantically search chunked PDF literature related to the process of interest. It retrieves both textual descriptions of parametric relationships and any available equations.

**TABLE 11:** Summarized Query Form (Knowledge Retrieval from Literature)

| Component | Description |
|---|---|
| Input | PDF documents related to the manufacturing process of interest; input process variables; output process variable to model |
| Output | Identified relationships between each input variable and the output; relationship types (e.g., linear, nonlinear, logarithmic); any relevant equations in the form output = f(inputs); supporting evidence from tables or text |
| Instructions | For each input variable: analyze its relationship with the output; if a relationship is found, record its type (positive, negative, logarithmic, polynomial, etc.) and retrieve the quantitative influence and supporting evidence.<br>If equations are present in the literature: extract and report in the form output = f(inputs) with an explanation of each term.<br>Include explanations of the process and variables. |

The prompt is submitted to the LLM (GPT-4o-mini) with the RAG-retrieved response. In the initial round it generates candidate equations; in subsequent refinement rounds the top 20 prior models and their performance metrics are included to guide improvement.

**TABLE 12** Summarized Prompt Form (Analytical Model Generation and Refinement)

| Component | Description |
| --- | --- |
| Input | Retrieved information (RAG response) describing parametric relationships; For refinement step, a summary of the top 20 best-performing models from previous iterations ($R^2$ and MSE scores) |
| Output | A Python function of the form:<br>def model (inputs, a0, ...):<br>input_1, input_2, ... = inputs<br>output = f (input_1, input_2, ...)<br>return output<br>mapping input parameters to the output parameter; no additional explanation or context. |
| Instructions (Initial Generation) | Use the retrieved relationships to generate a Python function modeling the output as a function of the inputs. If an equation was extracted from the literature, format it directly as the Python function. Otherwise, construct a function consistent with the described relationships, using appropriate function types (linear, exponential, logarithmic, trigonometric, etc.). Provide no other context or explanation, just the Python function code. |
| Instructions (Iterative Refinement) | Given the top 20 previously generated models and their $R^2$ / MSE scores; prioritize models with higher $R^2$ and lower MSE. Generate an improved function by applying algebraic manipulation, combining or introducing new terms, or modifying relationships between input and output parameters, guided by both the retrieved literature information and the best-performing prior equations. Output only the updated function. |